\documentclass[10pt,twocolumn,letterpaper]{article}

\usepackage{iccv}
\usepackage{times}
\usepackage{epsfig}
\usepackage{graphicx}
\usepackage{amsmath}
\usepackage{amssymb}

\usepackage[pagebackref=true,breaklinks=true,letterpaper=true,colorlinks,bookmarks=false]{hyperref}

\iccvfinalcopy 

\def\iccvPaperID{9629} 

\ificcvfinal\fi

\begin{document}

\title{Fast Non-line-of-sight Imaging with Two-step Deep Remapping}

\author{Dayu Zhu\\
Georgia Institute of Technology\\
{\tt\small dzhu71@gatech.edu}
\and
Wenshan Cai\\
Georgia Institute of Technology\\
{\tt\small wcai@gatech.edu}
}

\maketitle
\ificcvfinal\thispagestyle{empty}\fi

\begin{abstract}
   Conventional imaging only records photons directly sent from the object to the detector, while non-line-of-sight (NLOS) imaging takes the indirect light into account. Most NLOS solutions employ a transient scanning process, followed by a physical based algorithm to reconstruct the NLOS scenes. However, the transient detection requires sophisticated apparatus, with long scanning time and low robustness to ambient environment, and the reconstruction algorithms are typically time-consuming and computationally expensive. Here we propose a new NLOS solution to address the above defects, with innovations on both equipment and algorithm. We apply inexpensive commercial Lidar for detection, with much higher scanning speed and better compatibility to real-world imaging. Our reconstruction framework is deep learning based, with a generative two-step remapping strategy to guarantee high reconstruction fidelity. The overall detection and reconstruction process allows for millisecond responses, with reconstruction precision of millimeter level. We have experimentally tested the proposed solution on both synthetic and real objects, and further demonstrated our method to be applicable to full-color NLOS imaging. 
\end{abstract}

\section{Introduction}

Imaging is a seemingly mature technique that is employed to record the real-world scene, which has been developed over centuries. One central technique of imaging is to record the distribution of photons which are emitted from or scattered by the target and further received by the camera. Apart from the planar retrievals, i.e. pictures, people also strive to extract the depth or phase information of objects and reconstruct three-dimensional scenes, which is a thriving field of research in computer vision~\cite{3D_vision_saxena2005learning,3D_vision_book_cyganek2011introduction}. Most imaging and reconstruction methods can only recover the information in the line-of-sight (LOS) field of the camera, which implies that there should be no obstacle between the direct light path connecting the object and the camera. Otherwise, the light from the object will be reflected or deflected, which alters the intensity and directionality of light and further inevitably impedes and encrypts the original photon distribution. As a result, the object is ‘hidden’ from the plain sight and becomes non-line-of-sight (NLOS) for observation, and the encrypted information is usually too weak and has to be treated as noise in imaging. 
\begin{figure}[t]
\begin{center}
 \includegraphics[width=0.95\linewidth]{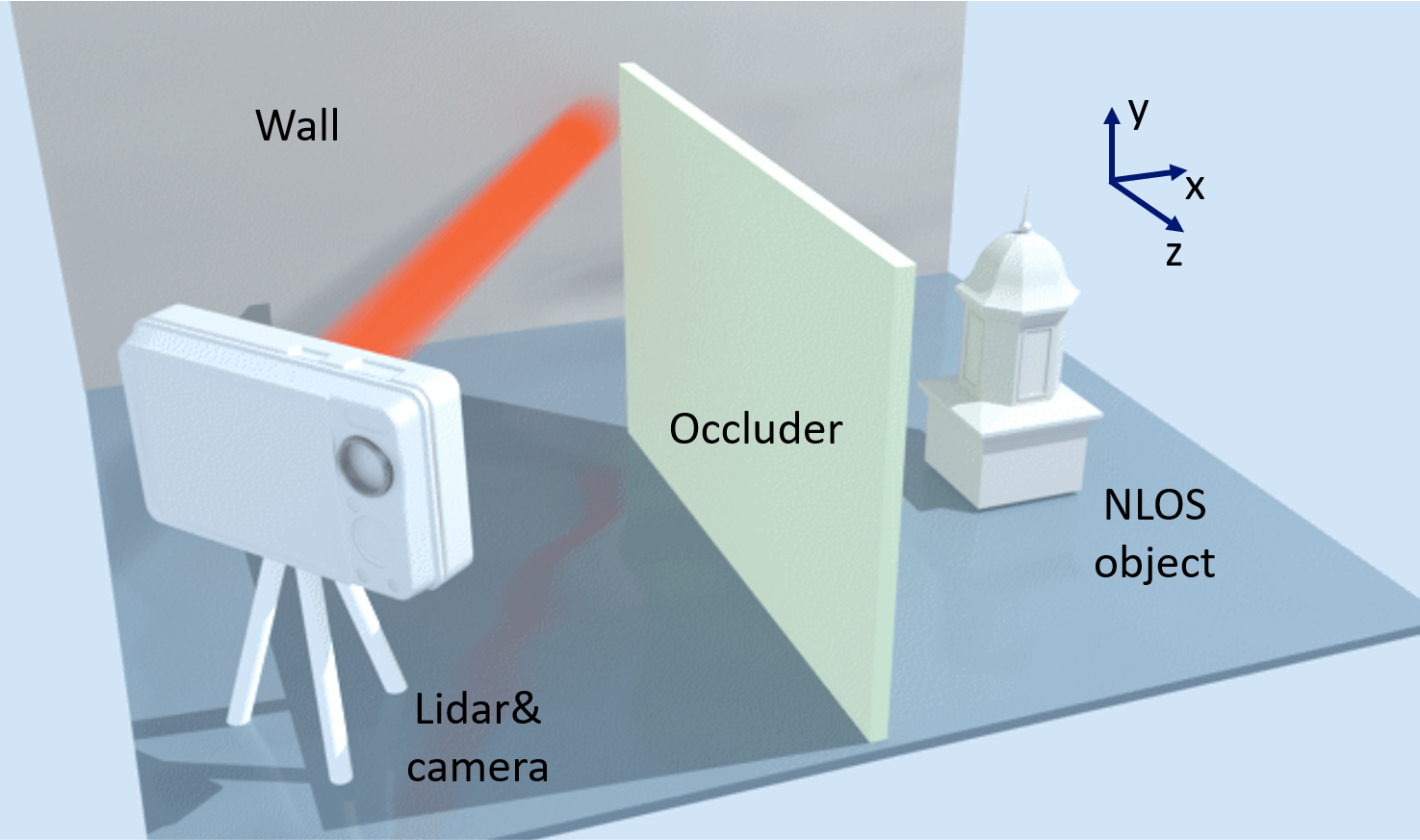}
\end{center}
   \caption{Illustrator of the non-line-of-sight (NLOS) imaging. The light path between the object and the camera is blocked by an occluder, thus the camera cannot directly capture the information of the object. With the aid of a wall or a secondary object which scatters the photons from the NLOS object to the camera, NLOS imaging may be achieved. Lidar is to actively probe the NLOS scene, and information of the reflected light is used to reconstruct the NLOS object.}
\label{fig:long}
\label{fig:1}
\end{figure}

Although the NLOS information is greatly distorted, it is still possible to recover the original photon distribution and achieve NLOS imaging and reconstruction~\cite{naturereview2020}. Numerous attempts have been made to analytically decode the NLOS information from the captured pictures~\cite{bouman2017_cornercamera,aittala2019computationalmirror,passive_batarseh2018passive_NC,passive_Baradad_2018_CVPR,incoherent_bertolotti2012noninvasive, incoherent_boger2019}. Despite effective in some cases, the LOS and NLOS data are greatly entangled on the image plane and there is always no guarantee of reasonable separation of them. Thus, instead of passively decoding the images, an alternative approach is to actively probe the NLOS scene ~\cite{velten2012recovering, liu2019non, xin2019theory,gariepy2016detection_tracking,o2018confocal,tsai2017geometry,active_heide2014diffuse,ahn2019convolutional_opticalway,active_thrampoulidis2018exploiting,sino2020}. This active method provides much more information than a passive image and much higher degree of freedom for NLOS imaging and reconstruction. This genre usually applies a laser to scan the scene spatially, and a photon detector is used to detect the transient events of back-scattered photons. The NLOS reconstruction is then achieved by analyzing the back-projection conditions of the transients or through other geometric optical transformations~\cite{backprojection_arellano2017fast,velten2012recovering}. The algorithms may proceed iteratively until the reconstruction converges. The active scheme can achieve high fidelity of NLOS reconstruction, meanwhile, some major shortcomings hinder the universal application of this NLOS imaging approach. On the physical level, laser scanning and transient detection processes are necessary. The laser scanning process may take minutes to hours with a lab-built scanning system, which is not applicable to real-time scenarios, and the ultrafast lasers are expensive and may not satisfy the safe standard for out-of-lab applications. As for the detection equipment, the transient events require lab-built optical systems with sophisticated apparatus, such as single-photon avalanche photodetectors (SPAD), ultrafast photon counting modules, streak cameras, etc~\cite{ghioni2007progress_spad, becker2005_photoncounting, itatani2002streakcamera}. The overall equipment settings are highly complicated and vulnerable to ambient perturbations, thus they are more suitable for ultrafast optical experiments than real-world imaging scenes. At the algorithm level, the transient detection will collect a huge amount of data, and the reconstruction process always deals with matrices or tensors of gigantic sizes. These properties lead to high consumption of memory, storage and computational resources, and the reconstruction time will hardly fulfill the expectation of real-time NLOS imaging and reconstruction. 

To address these pressing challenges, in this paper we introduce a novel methodology for real time NLOS imaging. Instead of using sophisticated devices for transient experiment, we employ a low-end Lidar as both the probe and detection devices~\cite{weitkamp2006lidar}. Compared to in-lab scanning systems with ultrafast laser and SPAD, Lidar is commercially available at a much lower cost (only a few hundred dollars), with substantially faster operation speed (scanning time at the millisecond level) and strong robustness to environmental factors. Not relying on transient time-of-flight (ToF) information of photons, we utilize the depth map and intensity map collected by the Lidar to achieve real-time, full-color NLOS imaging. Our algorithm is deep learning based and consists of two neural networks: a compressor and a generator~\cite{lecun2015deeplearning}. There are some pioneering works that have introduced deep learning into NLOS imaging and revealed the promising future of deep learning assisted NLOS imaging ~\cite{chen2019steady,chopite2020deep,lei2019direct, metzler2020deep_optica, aittala2019computationalmirror}. The reconstruction speeds of deep learning approaches are inherently much faster than geometric physical-based methods, with better versatility and generalization abilities. The reported methods are mostly built on supervised learning and try to establish the direct mapping between the detected data and target NLOS scenes. However, in sharp contrast to most computer vision tasks, in NLOS reconstruction the input data and output scenes share almost no common geometric features (such as edges or shapes), thus the supervised mapping of input and the label (i.e. the ground truth NLOS scene) may not be well-learnt by a conventional convolutional neural network (CNN)~\cite{lecun1998gradient, krizhevsky2017imagenet}. Besides, generating high dimensional data with explicit semantics, such as images or 3D scenes, is never a closed-form problem, and a generative model such as generative adversarial network (GAN) or variational autoencoder (VAE) is necessary~\cite{goodfellow2014GAN,kingma2013vae}. Therefore, the reconstruction performance of these reported methods is limited by the direct mapping with supervised learning, such as blurriness of reconstructed results and false retrieval of complicated objects. In this work we propose a deep learning framework that consists of two components and achieves NLOS reconstruction in two steps. The generator is the decoder of a trained VAE responsible for generating various scenes, and the CNN compressor converts the detected depth and intensity information into a low-dimensional latent space for the generator. When the Lidar collects new depth and intensity maps, the compressor will compress them into a latent vector, then the generator will recover the NLOS scene by decoding the vector. This methodology overcomes the flaws of previous deep learning approaches, which were based on a single network and trained in a supervised fashion. Our framework is capable of achieving state-of-the-art reconstruction performance, with strong robustness to ambient surroundings and the position, gesture and viewpoint of the NLOS objects. We have demonstrated the functionality of our approach on a synthetic dataset, then applied transfer learning and verified the performance through real-world experiments. To illustrate that our approach is not limited to the reconstruction of geometric features of NLOS scenes, we step further to experimentally prove the effectiveness of our method for the recovery of full-color information. The methodology presented here potentially offers a paradigm shift for applications of real-time NLOS imaging, such as driving assistant, geological prospecting, biomedical imaging, animal behavioral experiments, and many more~\cite{application_tracking_klein2016tracking,application_tracking_Smith_2018_CVPR,application_scheiner2020seeingstreet,gariepy2016detection_tracking,application_chan2017non,katz2014_medical,application_oe_xu2018revealing,application_pediredla2019snlos,application_realtime_2020real}.

\section{Methods}
\subsection{Lidar-based detection}
\begin{figure}[t]
\begin{center}
\includegraphics[width=0.95\linewidth]{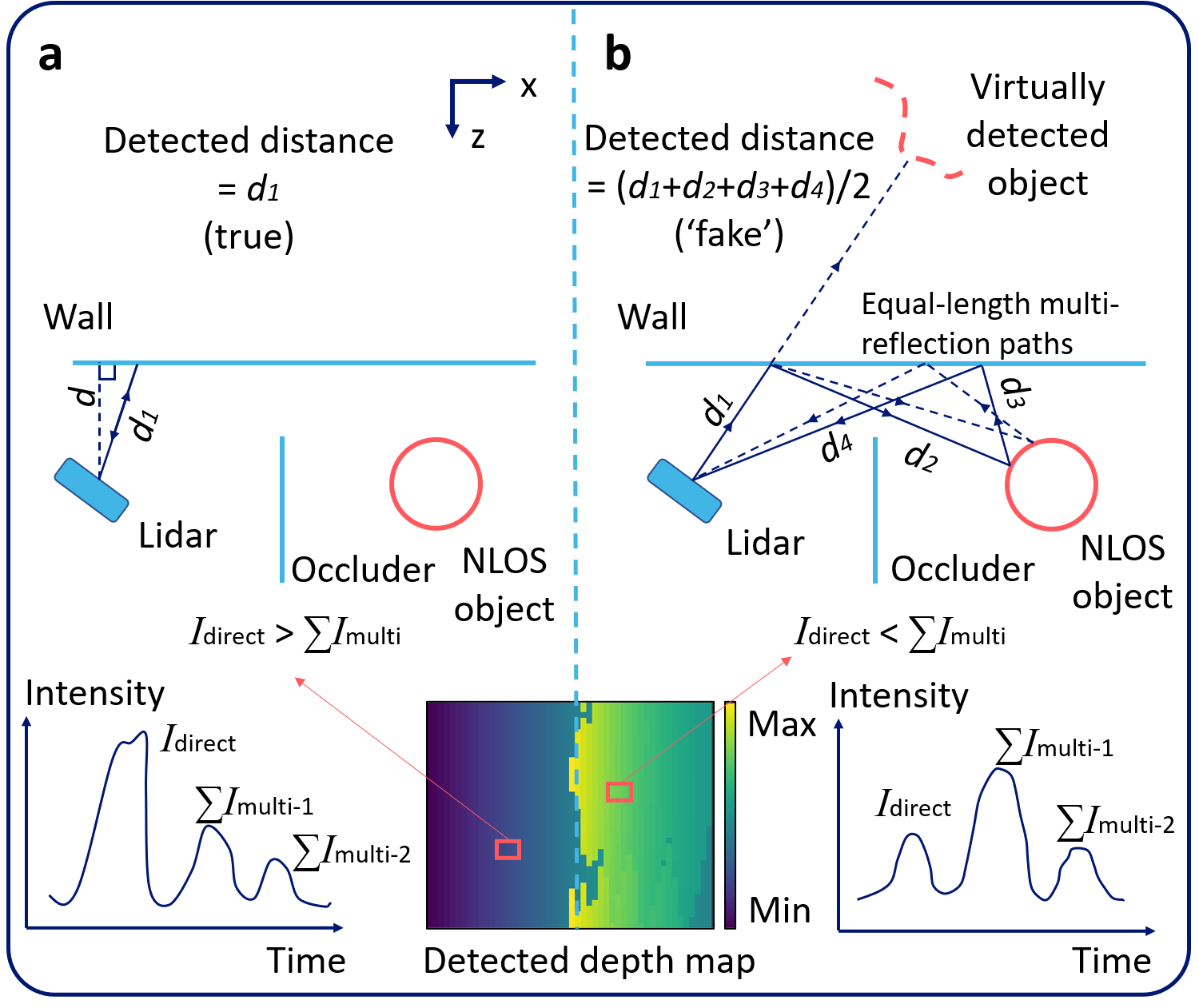}
\end{center}
   \caption{Two cases of Lidar detection (top view). (a) Generally, when the light directly reflected from the wall is far stronger than the light experiencing multi-reflection, the Lidar will denote the correct depth as the actual distance between itself and the wall. (b) If the total intensity of the beams undergoing equal-length multi-reflection is more intense than the direct-reflected light, the Lidar will store the intensity of multi-reflected light and treat the multi-reflection optical path length as the direct-reflection one, which leads to a ‘fake detected distance’.}
\label{fig:long}
\label{fig:2}
\end{figure}

The typical NLOS imaging scenario is depicted in Fig.~\ref{fig:1}. When an obstacle blocks the direct light path between the object and the camera, the conventional LOS imaging would fail to function as the detector receives no direct information from the target. However, another object or a wall between the camera and the target may serve as a ‘relay station’ and provide an indirect light path for NLOS imaging: a photon sent from the target object can be reflected by the relay object and further collected by the camera. After that, information from both the target and the relay objects is embedded in the photon distribution received by the camera. To disentangle the two sources of data, we need to actively probe the surroundings to extract data under different perturbations. The probe is usually conducted through sending a laser in the field of view (FOV) to scan the environment and then collecting the perturbed photon distributions. In this work we use a low-cost Lidar (Intel RealSense, the functioning details of the Lidar are provided in the Supplementary Material) as both the laser source and the camera. The laser beam on the Lidar scans the FOV of 70 by 55 degrees, then the detector on it sequentially collects the photons reflected from the surrounding objects. Based on the ToF information of the received photons, the on-chip processor of the Lidar will infer a depth map and a light intensity map. In general cases (Fig.~\ref{fig:2}a), most of the photons received by the Lidar are the ones directly reflected back and travel along the reverse path. There are also photons deflected by the wall to different directions, further scattered by surrounding objects and finally collected by the Lidar. However, the light intensities along these multi-reflection pathways will be much lower than that of the direct-reflection light. In this case, the Lidar will count the travel time of the direct-reflection photons and the calculated depth will represent the orthogonal distance between the Lidar plane and the incidence point on the wall, which is typical for the regular Lidar operation.

\begin{figure*}[t]
\begin{center}
\includegraphics[width=0.9\linewidth]{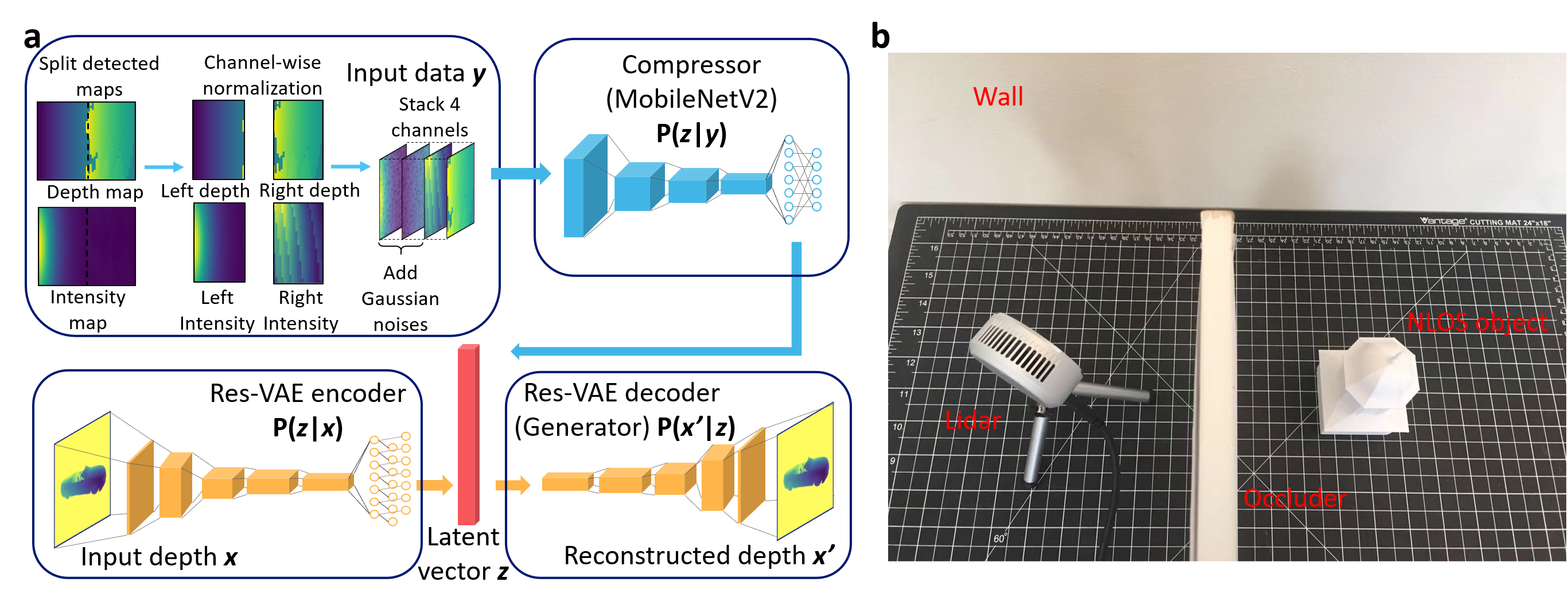}
\end{center}
   \caption{(a) Algorithm workflow and (b) experimental setup for NLOS imaging. (a) The NLOS reconstruction framework consists of two neural networks, a generator and a compressor. The reconstruction process is two-step: the compressor compresses the detected data as a latent vector, and the generator decodes the latent vector into a depth map of the NLOS object. The compressor is adapted from MobileNetV2, and the generator is the decoder of a variational autoencoder (VAE) with ResNet blocks. The input contains the depth and intensity maps detected by the Lidar, stacked as a 4-channel tensor with channel-wise normalization and Gaussian noises.  (b) The experimental setup for NLOS imaging used in this work. }
\label{fig:long}
\label{fig:3}
\end{figure*}

In another case (Fig.~\ref{fig:2}b), if the incidence angle is very oblique (\textgreater 45 degree) and the point of illumination on the wall is close to the target object (\textless 0.3 m in our cases), the multi-reflection paths may overshadow the direct-reflection path. Considering the three-bounce multi-reflection cases (Lidar-wall-object-wall-Lidar), there will be numerous possible three-bounce light paths with the same optical path length, and their total light intensity can be stronger than the direct-reflection one. As a result, the Lidar will denote the distance as half of the multi-reflection optical path length, i.e., the NLOS distance, rather than the direct distance. Although this so-called ‘multipath artifacts’ is undesired in conventional Lidar applications, it is favorable in our NLOS imaging ~\cite{multipath2012}. In the common case as shown in Fig.~\ref{fig:2}a, the Lidar will provide the correct depth and intensity, and the information represents the geometric and reflection features of the relay wall, which also implies the position of the Lidar relative to the wall. While in the NLOS case (Fig.~\ref{fig:2}b), even though the received photons by the Lidar contain information of actual light intensities and optical paths, the Lidar will incorrectly map them to ‘fake’ locations. In this way, the ground truth photon distribution that reveals the NLOS information will be distorted and overshadowed by the Lidar mechanism. Our methodology focuses on redistributing the intensity and depth maps to extract the original NLOS information and thus achieving imaging and reconstruction in a highly efficient and faithful manner. 

\subsection{Two-step deep remapping framework}
Since the redistributing or remapping function varies with different NLOS scenes, here we apply deep learning to learn the connection between the ground truth NLOS objects and detected maps of depth and intensity. There are some pioneering works that have introduced supervised learning and applied deep neural networks (NN, such as U-Net~\cite{ronneberger2015unet}) into NLOS imaging. However, the remapping task has two intrinsic requirements: (1) transform the collected information into NLOS representations, and (2) generate NLOS scenes with clean semantics ~\cite{chen2020_TOG}. As a result, a regression-oriented NN trained in the supervised fashion cannot simultaneously meet these two requirements with decent performance. In contrast, here we propose a reconstruction framework consisting of two networks, each fulfilling one requirement: a compressor transforms the detected depth and intensity maps into a latent vector, and a generator decodes the latent vector to the predicted NLOS scene. With the sequential functions of the two components, the algorithm can reconstruct NLOS scenes with high fidelity, comparable to the performance of the state-of-the-art transient-based methods, with a processing time of no more than milliseconds.  

The pipeline of the framework is depicted in Fig.~\ref{fig:3}a. The generator is the decoder of a VAE. Similar to GAN, VAE is a powerful tool for scene and image synthesis, and offers better controllability on the sampling of the latent space. A typical CNN-based VAE consists of an encoder and a decoder with mirror structures to each other. The VAE applied here is composed of ResNet blocks to improve the resolving power and training effectiveness, and we denote it as Res-VAE~\cite{he2016deep_resnet}. Specifically, the encoder down-samples an input image $x$ (64 by 64 in our experiments) to a 256-dimension latent vector $z$, which means the encoder learns the posterior $p(z|x)$, and $z$ is reparameterized to force its distribution $p(z)$ to be a standard Gaussian distribution. Further, the decoder ($p(x’|z)$) expands $z$ to a reconstructed image $x^\prime$. The training process is to lower the difference between $x$ and $x^\prime$ (reconstruction loss), as well as the distance between $p(z)$ and a standard Gaussian distribution (Kullback-Leibler divergence). Once the training is finished, the decoder will be able to generate images similar to the ones in the training dataset, and the latent vectors can be randomly sampled from the standard Gaussian distribution. In our case, the input $x$ is the depth map of the NLOS scene relative to the wall, which corresponds to the perpendicular distances between the wall and the elements on the point clouds of NLOS objects. Since the mapping between $x$ and $z$ is mostly one-to-one, the reconstruction is no longer necessary to transform the detected information $y$ to a high-dimension $x$. Instead, it only needs to compress $y$ to a low dimensional vector $z$, which can be facilitated by a CNN-based compressor. The role of the compressor is rather standard in computer vision, which is fed with images and generates the predicted labels. Here the compressor is adapted from a lightweight network, MobileNetV2, to achieve good performance and avoid overfitting~\cite{sandler2018mobilenetv2}. The input of the compressor is the depth-and-intensity map detected by the Lidar, and the output predicts the latent vectors of the target NLOS depth map encoded by the pretrained Res-VAE.

\section{Experiments}
To validate the efficacy of the Lidar-based approach and the deep learning reconstruction algorithm, we have conducted 4 categories of experiments. The first one is based on a synthetic dataset, and the second experiment is on real objects with transfer learning applied. We further demonstrate the power of our method for full-color NLOS imaging, with experiments on images of everyday objects and complex scenes, respectively.

\begin{figure*}[t]
\begin{center}
\includegraphics[width=0.8\linewidth]{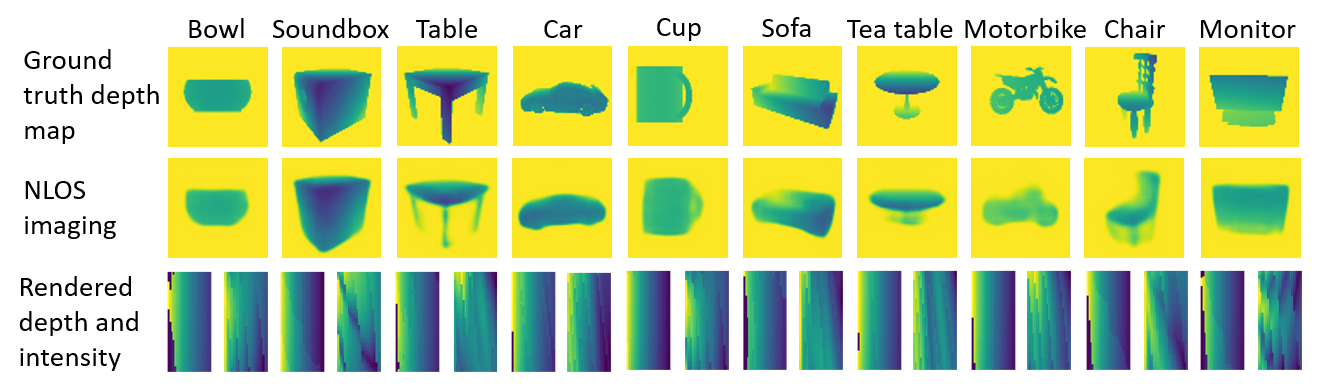}
\end{center}
   \caption{The NLOS reconstruction experiment on a synthetic dataset. The reconstruction results of randomly selected everyday items in the test set are presented. The first row is the ground truth depth maps of the items, and the second row shows the reconstructed depth maps. The last row presents the inputs for the NLOS reconstruction, which are the simulated depth and intensity maps by our NLOS renderer. Since we assume the NLOS objects are situated on the right of the Lidar, the left halves of the depth and intensity maps are almost identical for different items. Therefore, here we only present the right halves of these maps. For each case, the left image represents the depth map, and the one on the right is for the intensity.}
\label{fig:long}
\label{fig:4}
\end{figure*}

\subsection{Experiment on a synthetic dataset}
\noindent {\bf NLOS renderer}. We first train and test the functionality of our framework on a synthetic dataset, and the weights can be further optimized when applied to real-world experiments through transfer learning. The synthetic dataset is composed of the NLOS depth maps and the corresponding simulated depth and intensity maps detected by the Lidar. As a result, a NLOS renderer is necessary to simulate the behavior of the Lidar~\cite{rendering_jarabo2014framework}. Our rendering framework is based on three-bounce simulation. For each incidence, if there is one multi-reflection optical path length whose total intensity (total intensity is the intensity summation of all the three-bounce light paths with the same optical path length) is greater than the direct-reflection intensity, the NLOS depth and intensity will be denoted. Otherwise, it will be a common LOS case. The implementation details of the renderer are described in the Supplementary Material. The final depth intensity maps consist of 80 by 64 pixels. Since the renderer conducts matrix operations and multiprocessing calculation, the rendering for one object takes only seconds, which is much more efficient than traditional ray-tracing renderers. The targets for rendering are around 30,800 depth maps of everyday objects (cars, tables, lamps, boats, chairs, etc.) from ShapeNet~\cite{chang2015shapenet,NEURIPS2019_shapenet_renderer}. To improve the variance of the dataset, the objects are situated with various altitudes, angles and distances to the wall.

\begin{figure*}[t]
\begin{center}
\includegraphics[width=0.8\linewidth]{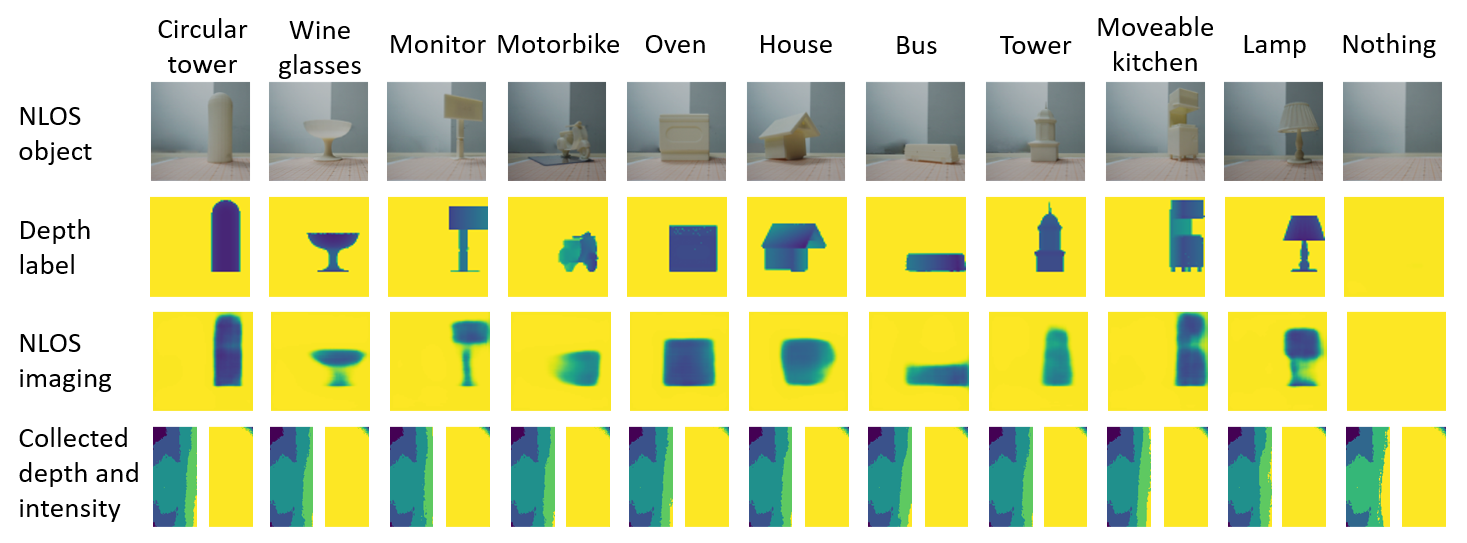}
\end{center}
   \caption{The NLOS reconstruction experiment on real objects. The first row displays the pictures of the actual objects used in this experiment, and the second row corresponds to the depth labels for training. The reconstructed depth maps are shown in the third row. The last row presents the inputs for the NLOS reconstruction, which are the depth and intensity maps collected by the Lidar (right halves only as explained before). }
\label{fig:long}
\label{fig:5}
\end{figure*}

\noindent {\bf Networks training.} Once the dataset is formed, it is time to train the neural networks in our framework. The Res-VAE needs to be trained first since we need to push the corresponding latent vectors of the depth maps into a dictionary. During the training of Res-VAE, the Kullback-Leibler distance is not necessary to be suppressed to a very low level. Since the compressor will infer the prior distribution $p(z)$ through $p(z|y)$, it is insignificant if $p(z)$ is not strictly Gaussian. With the acquisition of the latent vectors, the compressor is trained to map the rendered depth and intensity maps to the latent vectors. As illustrated in Figs.~\ref{fig:1} and~\ref{fig:3}b, we assume the Lidar is located on the left of the object, so the left halves of the depth and intensity maps will mostly denote the LOS information of the wall, while the right parts are encoded with the information of both the wall and the NLOS objects. The right halves of the depth maps correspond to larger distances, thereby leading to weaker intensities. To prevent the subtle signals on the right from being flooded by the overwhelming information on the left, we split the maps to the left and right halves, and vertically stack them into a 4-channel input image (ordered as right-depth, right-intensity, left-depth, left-intensity), with each channel normalized independently as illustrated in Fig.~\ref{fig:3}(a). Besides, the left-depth and left-intensity are almost the same for different data points since they mostly document the LOS information. Hence, we introduce Gaussian noises to these two channels in order to add randomness and prevent overfitting. The 30,800 data points are randomly split into a training dataset of 24,600 and a test dataset of 6,200. The details of the training process and the structures of the generator and the compressor are provided in the Supplementary Material.  

\noindent {\bf Reconstruction performance.} With the training of both the generator and the compressor, we will be able to predict the NLOS scenes based on the captured depth and intensity maps. The NLOS reconstruction results of different categories of ShapeNet-items from the test dataset are plotted in Fig.~\ref{fig:4}. The first two rows are the ground truth depth maps and the reconstructed results, respectively. The third row of the figure presents the rendered depth and intensity maps which are used as the input for the reconstruction framework. Since their left halves are almost identical and indistinguishable to the eye, we only show their right halves for each case. Although the examples are randomly chosen and with a wide range of shapes, the reconstruction algorithm is able to recover significant features of them. This experiment demonstrates the high quality of the NLOS imaging through the developed method in this work. 

\subsection{Real-world NLOS reconstruction}
Having confirmed the performance of our NLOS imaging method on the synthetic dataset, we further apply it to real-world NLOS scenes. In real-world applications we usually cannot conduct an enormous amount of detections to generate a big training dataset, therefore we utilize the networks with weights pretrained on the synthetic dataset, further continue to train the networks on a relatively small real-world dataset through transfer learning. We have 54 3D-printed objects as the target objects, which are models of common objects like buses, tables, pianos, etc. The experimental environment is presented in Fig.~\ref{fig:3}b, with the objects positioned at 9 different locations with 5 rotational angles (rotation only for non-circular items), generating a dataset of around 1,900 NLOS cases. Although the generator does not need to retrain, the compressor trained on the synthetic dataset is not directly applicable to the real-world environment. Besides, such a tiny dataset (randomly split into 1,600 as the training dataset and 300 as the test set) is not compatible to a large-scale network due to inevitable overfitting, and a simple network is unable to handle the task. In this case, we leverage transfer learning to mitigate the training dilemma and transfer the knowledge that the compressor has learnt from the synthetic dataset to the real-world dataset. The target labels for the compressor are the latent vectors of the virtual depth maps corresponding to the 1,900 scenes. Since the compressor has been previously trained on the synthetic dataset, we continue to train it on the real-world training dataset for several epochs until the training loss and test loss are about to diverge. Next, we freeze all the network weights except for the last linear layer, and train it until the loss becomes steady. The entire process of the transfer learning takes only a couple of minutes, and the reconstruction performance on the test dataset is displayed in Fig.~\ref{fig:5}. Each column presents one example, and the four rows are the pictures of the actual objects, the virtual depth labels, the reconstructed depth maps, and the right halves of detected depth and intensity maps, respectively. It is worth noting that our methodology is able to reconstruct sufficient geometric and positional information of the NLOS objects. And if there is nothing behind the occluder, the framework will not provide a false-positive NLOS result. Certain details are not fully recovered (such as the motorbike), which is expected since some of the reflected light cannot be captured by the Lidar, thus the loss of information is inevitable. Nevertheless, the reconstruction quality is comparable or superior to those of the state-of-the-art methods, with much lower requirements on equipment and ambient environment along with much faster imaging speed. Since the imaging scale and data format are all different from other methods, it is impossible to compare them quantitatively, while a qualitative comparison and error analysis are provided in Supplementary Material. The average accuracy of the reconstructed depth is 98.27$\%$ (root mean square error of 4.68 mm, see Supplementary Material for metric definition). Such a performance indicates our methodology is applicable to small-scale NLOS scenarios that require good precision and fast speed, such as biomedical imaging, animal behavioral surveillance, etc. If we further use open-source Lidar that allows access to detailed ToF information, the NLOS signal will be captured even though it is no stronger than the LOS signal. In this case, the imaging will not be limited to the cases where the multipath artifacts happen. Besides, the maps acquired by the Lidar will indicate the position of Lidar and bidirectional reflectance distribution function (BRDF) of the wall, which functions for calibration and makes the model readily transferable to other imaging setups (such as positional displacement of Lidar, material change of wall or objects) by meta-learning or phase compensation. As all deep learning algorithms are data-driven, we will enlarge our dataset in further study, and setup-independent NLOS imaging will also be formalized in future. Still, since the model has learnt the inherent physics of photon remapping, it will be able to recover some NLOS features even if the imaging condition is altered or the target is far from the distribution of the dataset. 

\subsection{Full-color NLOS imaging}
NLOS reconstruction mostly refers to the reconstruction of the geometric features of NLOS objects, while regular LOS imaging also captures and records the color information. Here we step forward to demonstrate the capacity of our methodology for the recovery of NLOS scenes with full color information. 

\begin{figure}[t]
\begin{center}
 \includegraphics[width=0.85\linewidth]{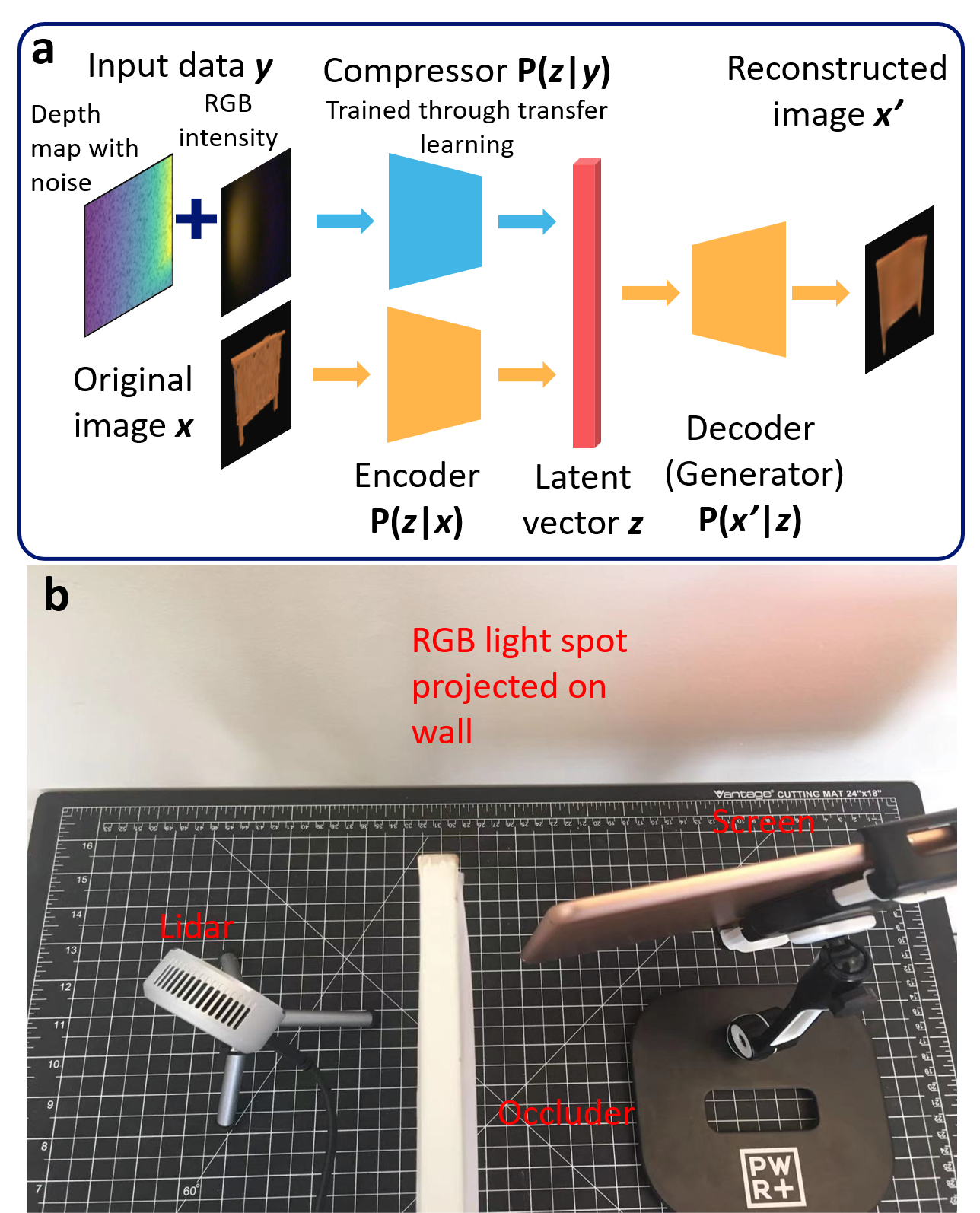}
\end{center}
   \caption{(a) Algorithm workflow and (b) experimental setup for full-color NLOS imaging. (a) The algorithm is essentially the same as the one presented in Fig.~\ref{fig:3}, while the input here is the concatenated tensor of captured RGB light intensity map and the depth map with Gaussian noise. The reconstruction objective is to obtain the full-color image. (b) When the screen displays an image, the device is able to detect the depth map along with the RGB light intensity map projected on the wall, which will be used for full-color NLOS imaging.}
\label{fig:long}
\label{fig:6}
\end{figure}

\begin{figure*}[t]
\begin{center}
\includegraphics[width=0.75\linewidth]{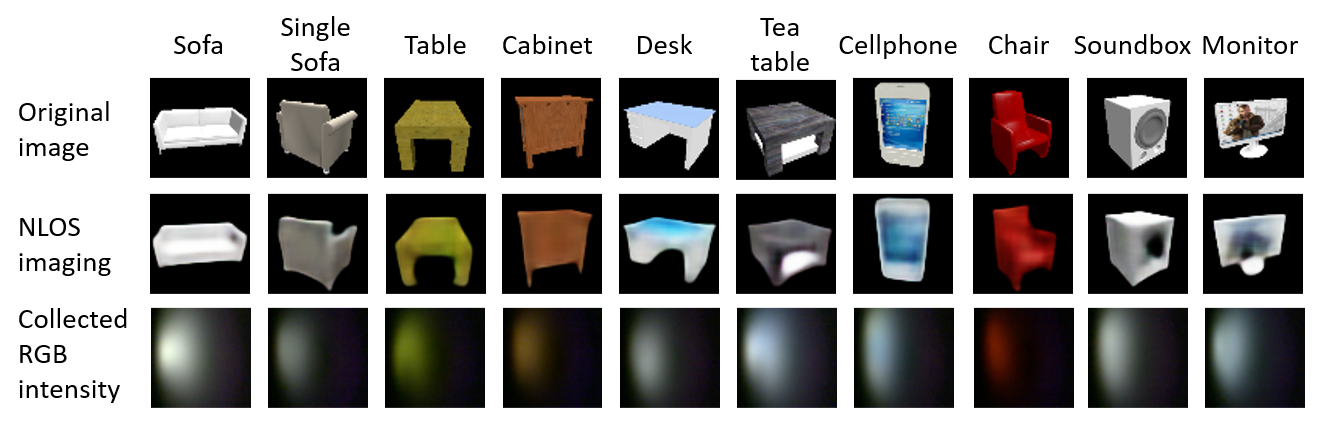}
\end{center}
   \caption{Full-color NLOS reconstruction experiment on displayed images of everyday items. The results of different categories of everyday items are presented, with the first row as the original images, and the second row as the reconstructed full-color images. The third row presents the RGB light intensity maps captured by the camera on the Lidar.}
\label{fig:long}
\label{fig:7}
\end{figure*}

Most commercial Lidars utilize infrared laser to perceive the surrounding environment. However, it is impossible to detect the color information in the visible wavelength range through an infrared laser at a single wavelength. One promising solution is to build an optical system with red, green and blue (RGB) lasers as the scanning seed~\cite{chen2019steady}. While stick to commercially available devices, we demonstrate a Lidar and a camera are able to cooperatively recover the color images displayed in the NLOS scenes. Since the Lidar is equipped with a camera, we do not need to add additional detectors or components to the system. The experimental setup for full-color NLOS imaging is illustrated in Fig.~\ref{fig:6}. A screen displays images but is invisible to the camera, while we aim to recover the colored images based on the light projected to the wall. In this experiment, the reconstruction objective will not be single-channel depth maps, but three-channel RGB images instead~\cite{saunders2019getcolorimage}. The algorithm is similar to previous, as depicted in Fig.~\ref{fig:6}a. The training dataset is composed of RGB images of everyday objects. Each time the screen displays a new image, the camera will capture an image of the light spot on the wall, and the Lidar also denotes the depth map. We have collected 48,800 data points and 43,000 of them are grouped into a training set. During the training phase, the Res-VAE is trained to generate RGB images of the objects, and the latent vectors of the images in the training set are documented. As for the compressor, the input is a 4-channel tensor, which is stacked with the RGB image collected by the camera along with the depth map detected by the Lidar. Here we add random Gaussian noises to the depth maps to introduce variance. The depth maps will function more significantly if the angle and position of the Lidar are dynamic, which is left to future research. To illustrate that the functionality of our NLOS framework is not limited to specific neural networks, here the compressor is adapted from ResNet50, which is also more compatible to the large-scale dataset with RGB information. The reconstructed RGB images of different categories of everyday items are displayed in Fig.~\ref{fig:7}, with original images, reconstructed images, and RGB light intensity maps arranged from the top row to the bottom. It is evidenced that the geometric features and color information are retained to a great extent, which validates the capacity of the proposed approach. Providing the RGB scanning light source is introduced, we will be able to recover both the shape and color of nonluminous objects.

\begin{figure}[t]
\begin{center}
 \includegraphics[width=0.8\linewidth]{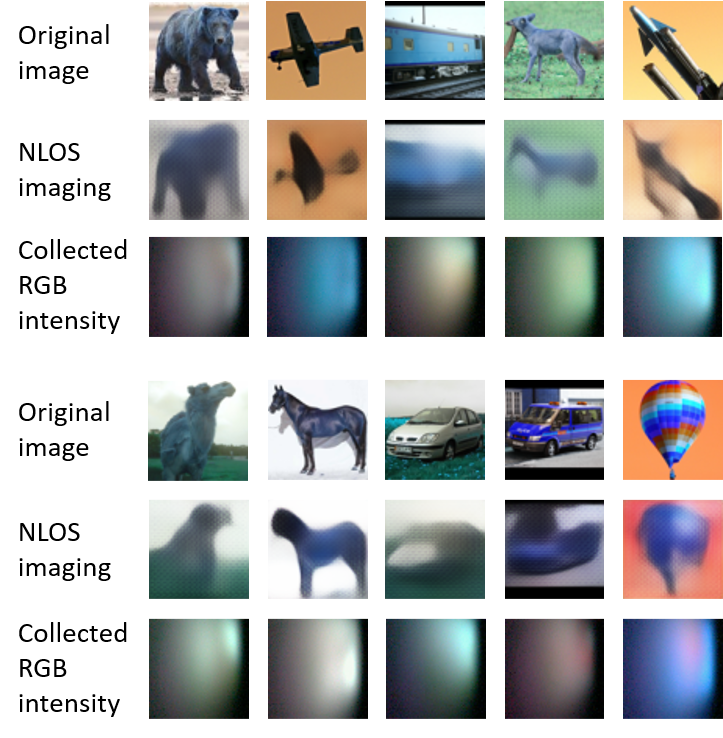}
\end{center}
   \caption{Full-color NLOS reconstruction experiment on displayed images of complex scenes. Randomly selected examples from the whole test set are presented, with the original images, the reconstructed full-color images, and the RGB light intensity maps shown sequentially. }
\label{fig:long}
\label{fig:8}
\end{figure}

\subsection{Full-color NLOS imaging for complex scenes}
A mature imaging technique should not have constraints on the number of objects to be captured, and the imaging quality should not degrade with the increased complexity of the scene. To explore the full capacity of our solution, we test it on a much more complicated RGB dataset, STL10~\cite{stl10}. Instead of images of isolated objects in pure black background, STL10 contains real pictures taken with different devices, angles, positions, and surrounding environments. The pictures may have more than one objects, and the semantic meanings are intricate and entangled with each other. To accommodate the increased amount of information in the image, we extend the dimension of the latent vector from 256 to 512, while other parameters and the training pipeline remain the same as those in the previous experiments. Some examples from the test set (50,000 data points in the training set, 10,000 in the test set) are presented in Fig.~\ref{fig:8}. Since the images in STL10 are not well-categorized, the examples are randomly picked from the whole test set. It is noticeable that major profiles and color information are correctly unfolded, while some fine details are sacrificed. The performance is mainly limited by the generative capability of the generator. Introducing advanced VAE models, such as $\beta$-VAE, IntroVAE, VQ-VAE-2, is expected to mitigate this issue~\cite{burgess2018beta-vae,huang2018introvae,razavi2019vqvae2}.            

\section{Conclusion and discussion}
We have developed a general methodology for real-time NLOS imaging, with innovation in both imaging equipment and reconstruction algorithm. Physically, we choose low-cost Lidar to replace the complicated devices commonly used for ultrafast experiment, and our scheme features much faster detection speed and better robustness to ambient environment. Algorithmically, we have proposed a deep learning framework consisting of two networks, a compressor and a generator. The framework is not in the typical manner of supervised learning, and the reconstruction process is two-step. The efficacy of the methodology has been verified experimentally, on both real-object NLOS reconstruction and full-color NLOS imaging, with the state-of-the-art performance. Our approach is directly pertinent to some real-world applications, such as corner-detection for unmanned vehicles, NLOS remote sensing, NLOS medical imaging, to name a few.

We realize additional efforts are needed to bring the proposed method to perfection. The Lidar used in this work has a very low laser power, which is designed for small range detection. We plan to transfer our technique to high-performance Lidar with open-source access to ToF information, such as those equipped on self-driving cars, to fulfill large-scale NLOS tasks, including but not limited to NLOS imaging of street scenes, buildings, and vehicles. Other follow-up efforts include the introduction of RGB light sources for full-color detection of nonluminous objects, development of techniques to disentangle the effects of the position and angle of Lidar, etc. As for the algorithm, the projected bottleneck for complex NLOS tasks will be the generative capability of the generator. Compared to GANs, the generated images of most VAEs are blurrier. Introducing most advanced VAE architectures is expected to resolve this problem. Since the compressor is responsible for extracting NLOS information from the depth and intensity maps, it would be favorable to update attention blocks to energize the compressor with better efficiency on locating the NLOS features. In addition, instead of training the generator and the compressor sequentially as performed in this work, we expect improved performance if they are trained concurrently, similar to the mechanism of GAN.

\section{Acknowledgements}
We thanks Su Li from Taxes A\&M University and Jingzhi Hu from Peking University for fruitful discussions.

{\small
\bibliographystyle{ieee_fullname}
\bibliography{egbib}
}

\end{document}